%% file: egpaper_final.tex
\documentclass[10pt,twocolumn,letterpaper]{article}

\usepackage{iccv}
\makeatletter
\@namedef{ver@everyshi.sty}{}
\makeatother
\usepackage{tikz}
\usepackage{times}
\usepackage{epsfig}
\usepackage{graphicx}
\usepackage{amsmath}
\usepackage{amssymb}
\usepackage{booktabs}
\usepackage{multirow}
\usepackage{multicol}
\usepackage{makecell}
\usepackage{soul}
\usepackage{pifont}
\usepackage[symbol]{footmisc}

\usepackage{url}
\usepackage{xcolor}
\usepackage{bbding}
\usepackage{utfsym}

\newcommand{\vb}[1]{\mathbf{#1}}
\newcommand{\vf}{\mathbf{f}}
\newcommand{\vR}{\mathbf{R}}
\newcommand{\vt}{\mathbf{t}}
\newcommand{\dlt}[1]{}

\newcommand{\nerffunc}{F_{\theta}}
\newcommand{\density}{\sigma}
\newcommand{\coord}{\mathbf{x}}
\newcommand{\pcolor}{\mathbf{c}}
\newcommand{\viewdir}{\mathbf{d}}
\newcommand{\ray}{\mathbf{r}}
\newcommand{\rcolor}{\mathbf{C}}

\usepackage[breaklinks=true,bookmarks=false]{hyperref}

\iccvfinalcopy 

\def\iccvPaperID{****} 

\ificcvfinal\fi

\begin{document}

\title{MovingParts: Motion-based 3D Part Discovery in Dynamic Radiance Field}

\author{Kaizhi Yang\textsuperscript{1}
\quad Xiaoshuai Zhang\textsuperscript{2}
\quad Zhiao Huang\textsuperscript{2}
\quad Xuejin Chen\textsuperscript{1}
\quad Zexiang Xu\textsuperscript{3\footnotemark[1]}
\quad Hao Su\textsuperscript{2\footnotemark[1]}
\\
\textsuperscript{1} University of Science and Technology of China \quad 
\textsuperscript{2} University of California, San Diego \\
\textsuperscript{3} Adobe Research
\\
\tt\small ykz0923@mail.ustc.edu.cn \quad 
\tt\small \{xiz040, z2huang, haosu\}@eng.ucsd.edu \\
\tt\small xjchen99@ustc.edu.cn \quad 
\tt\small zexu@adobe.com }
\maketitle
\footnotetext{*equal advisory}

\begin{abstract}
We present MovingParts, a NeRF-based method for dynamic scene reconstruction and part discovery.
We consider motion as an important cue for identifying parts, that all particles on the same part share the common motion pattern. From the perspective of fluid simulation, existing deformation-based methods for dynamic NeRF can be seen as parameterizing the scene motion under the Eulerian view, i.e., focusing on specific locations in space through which the fluid flows as time passes. However, it is intractable to extract the motion of constituting objects or parts using the Eulerian view representation. In this work, we introduce the dual Lagrangian view and enforce representations under the Eulerian/Lagrangian views to be cycle-consistent. Under the Lagrangian view, we parameterize the scene motion by tracking the trajectory of particles on objects. The Lagrangian view makes it convenient to discover parts by factorizing the scene motion as a composition of part-level rigid motions. Experimentally, our method can achieve fast and high-quality dynamic scene reconstruction from even a single moving camera, and the induced part-based representation allows direct applications of part tracking, animation, 3D scene editing, etc.
\end{abstract}

\input{sections/intro}
\input{sections/relatedwork}
\input{sections/prelim}
\input{sections/method}
\input{sections/experiments}

\input{sections/discussion}

{\small
\bibliographystyle{ieee_fullname}
\bibliography{egbib}
}

\end{document}

%% file: sections/intro.tex
\section{Introduction}
\label{sec:intro}

\begin{figure}[h]
  \centering
  \includegraphics[width=\linewidth]{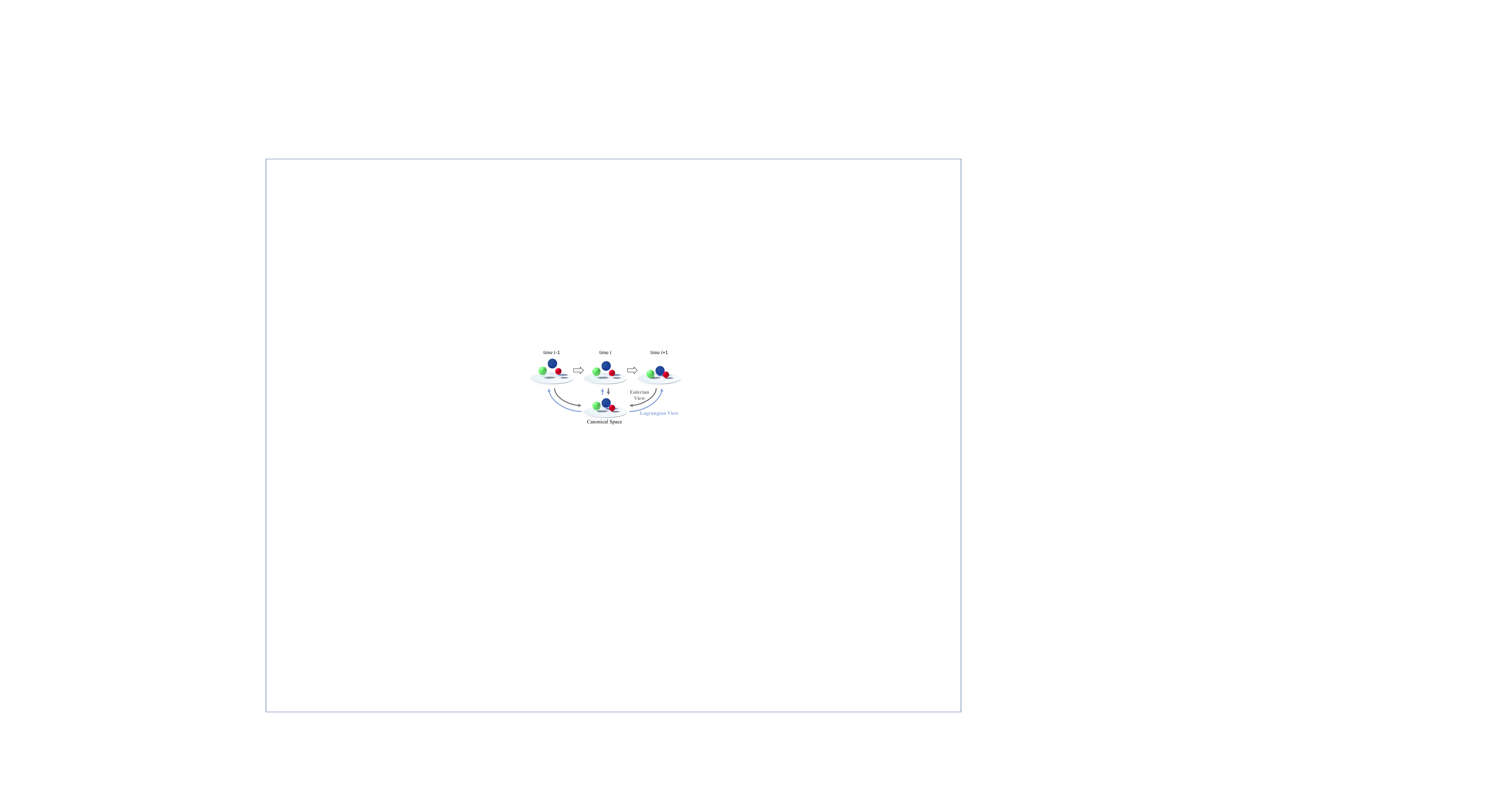}
  \caption{Location-based \emph{Eulerian view} vs. particle-based \emph{Lagrangian view}.
  The Eulerian view observes the flow at a specific location and the Lagrangian view observes the trajectory of specific particles. These two views constitute the conversion between the world space of each temporal frame and a canonical space.}
  \label{fig:EulerLag}
\end{figure}

3D scene reconstruction and understanding is one of the central problems in computer vision and graphics, with a wide range of applications in mixed reality, robotics, movie production, etc.
While many works focus on static scenes, real-world physical scenes are usually dynamic and entangled with illumination changes, object motion, and shape deformation.
The reconstruction of dynamic scenes is known to be highly challenging.
Nonrigid structure from motion methods~\cite{bregler2000recovering, gotardo2011non, kong2019deep, sidhu2020neural} could recover nonrigid shapes but are limited to sparse feature tracking.
To reduce the ambiguity between shape and motion, some other methods introduce multi-view capture~\cite{zhang-cvpr2003-ststereo, 10.1007/978-3-319-10593-2_3, 5459384} or category-specific priors~\cite{MorphableFace, livecap2019, kocabas2019vibe}.
Recently, neural radiance representation NeRF~\cite{mildenhall2020nerf} is applied to this field and achieved promising dynamic capture performance on general scenes using only monocular input~\cite{pumarola2020d, park2021nerfies, Li_2022_CVPR}.
However, most NeRF-based dynamic scene modeling methods only focus on scene reconstruction without considering scene understanding, thus lacking the ability to directly support downstream applications that need tracking, shape editing, re-animation, etc.

Our goal is to enable practical dynamic scene capture with both high-quality reconstruction and meaningful scene understanding from monocular input.
To this end, we propose MovingParts, a novel NeRF-based approach that can achieve not only fast dynamic scene reconstruction but also automatic rigid part discovery.
Our key insight is that motion (while complicating the reconstruction) can be used as an effective cue for identifying object parts because the scene content belonging to one rigid part has to share the same rigid transformation at any time during the capture.
Therefore, we design novel modules to explain the motions in dynamic neural fields, enabling unsupervised object part discovery via motion grouping. 


Our approach is inspired by the literature of fluid simulation. We note that a family of previous dynamic NeRF methods model scene motion using a 3D field that encodes scene flow~\cite{li2020neural} or deformation~\cite{pumarola2020d,park2021nerfies}. Specifically, at time $t$, for each location $\coord$, the 3D field encodes which particle $\coord_c$ in the canonical frame has been deformed to $\coord$. This is essentially the Eulerian view in fluid simulation~\cite{10.1145/383259.383260} -- motion information is denoted as a function $\Psi_E(\coord, t)$ at each specific location $\coord$ in the world coordinate frame.
It is known that, while the entire scene motion can be modeled under the Eulerian view, object/part motion at different temporal moments is actually intractable and hard to analyze. On the other hand, the Lagrangian view~\cite{10.1145/2601097.2601152}, as the duality of the Eulerian view, uses the particle-based representation to track the motion of each particle belonging to the object, and the constructed particle trajectory can be an important clue for scene analysis.
A relevant Lagrangian-based work is Watch-It-Move~\cite{noguchi2021watch} which composes objects into several ellipsoid-like parts by rendering supervision.
However, the multi-view requirement and the ellipsoidal geometric prior highly limit its application.
In contrast, we mainly focus on monocular input.


To achieve meaningful scene understanding by motion analysis, we propose a hybrid approach that learns motion under both the Eulerian and the Lagrangian views. 
In particular, our neural dynamic scene model consists of three modules: (1) a canonical module that models the scene geometry and appearance as a radiance field in a static canonical space, (2) an Eulerian module $\Psi_E(\coord, t)$ that records which particle $\coord_c$ in the canonical space passes through each specific location $\coord$ in the world coordinate frame at every time step, and (3) a Lagrangian module $\Psi_L(\coord_c, t)$ that records the trajectory of all particles $\coord_c$ in the canonical space.
Note that the motions modeled by the Eulerian and Lagrangian modules are inherently reciprocal, we, therefore, apply a cycle-consistency loss during reconstruction to enforce the consistency between the two modules, constraining them to model the same underlying motion in the scene. 

The construction of the Lagrangian view makes it convenient to discover parts by factorizing $\Psi_L(\coord_c, t)$. 
As the particles in a rigid part share a
common rigid transformation pattern, we propose a novel motion grouping module as part of our Lagrangian module.
By projecting the particle motion features into a few groups, we divide the scene into meaningful parts.
Once reconstructed, our Lagrangian module could offer part-level representation and 
allow for direct downstream applications such as part tracking, object control, and scene editing. 
Since the number of rigid parts generally differs across scenes, we introduce an additional group merging module that can adaptively merge the over-segmented groups into a reasonable number of rigid parts.




We jointly train all modules with only rendering supervision.
We demonstrate that our approach achieves high-quality dynamic scene reconstruction and realistic rendering results on par with state-of-the-art methods.
More importantly, compared with previous monocular NeRF methods, ours is the only one that simultaneously achieves part discovery, allowing for many more downstream applications.
Finally, inspired by recent fast NeRF reconstruction methods~\cite{Sun_2022_CVPR,chen2022tensorf,yu2021plenoxels}, we construct our system with feature volumes and light-weight multi-layer perceptrons (MLPs), leading to a fast reconstruction speed comparable to other concurrent methods that are specifically focused on speeding up dynamic NeRF. 

In summary, our key contributions are:
\begin{itemize}
\item We propose a novel NeRF-based method for simultaneous dynamic scene reconstruction and rigid part discovery from monocular image sequences;
\item The hybrid representation of feature volume and neural network allows us to achieve both high-quality reconstruction and reasonable part discovery within 30 minutes;
\item The extracted part-level representation can be directly applied to downstream applications like part tracking, object control, scene editing, etc.
\end{itemize}


%% file: sections/relatedwork.tex
\section{Related work}
\noindent\textbf{Dynamic Neural Radiance Fields.} 
Recently, the emergence of Neural Radiance Fields (NeRF)~\cite{mildenhall2020nerf} has facilitated the tasks of scene reconstruction and image synthesis.
Due to the dynamic properties of the physical world, an important branch of NeRF research is to extend it to dynamic scenes~\cite{pumarola2020d,li2020neural,park2021hypernerf}.
In particular, some methods directly extend the 5D radiance field function to 6D by adding additional time-dependent input to the network~\cite{Li_2022_CVPR, Xian_2021_CVPR}.
Other works enhance the temporal consistency in the 6D dynamic radiance field by explicitly modeling dynamic scene flows~\cite{li2020neural, du2021nerflow, Gao-ICCV-DynNeRF}, leading to promising results from only monocular input.
Meanwhile, deformation modules have also been adopted in NeRF-based methods~\cite{pumarola2020d, tretschk2021nonrigid, yuan2021star, park2021nerfies, park2021hypernerf, liu2022devrf}, offering strong regularization for temporal consistency.
Note that these various NeRF-based methods all explain motions (modeled as flows or deformation fields) from the location-based Eulerian view and do not support part discovery.
We instead propose a hybrid model that models motions with both location-based Eulerian and particle-based Lagrangian views, enabling high-quality dynamic scene reconstruction with automatic part discovery based on particle motion.

\noindent\textbf{Part Discovery from Motion.}
At the image level, most motion-based object discovery methods~\cite{bideauECCV16, yang2021selfsupervised, Xie_2019_CVPR, 6751331} cluster 2D pixels into different groups based on optical flow-related features.
Instead, we build our motion group module on the canonical 3D particles and rely on the predicted 3D rigid motion, which ensures arbitrary viewpoints consistency and temporal consistency of the grouping results.
In the 3D domain, some methods~\cite{Yaohao_2021,kawana2022unsupervised} reason about object parts by constructing point-wise correspondence at different object states and clustering their trajectories.
Without 3D input, \cite{8332511} adopts non-rigid structure from motion to reconstruct the 3D shape and applies spatio-temporal clustering to the 3D points to reason about segmentation.
However, only the geometry of sparse feature points could be achieved.
Recently, NeRF-based dynamic scene decoupling methods~\cite{yuan2021star, tschernezki21neuraldiff, wu2022dnerf} have been proposed.
Although they achieve dynamic scene decomposition with high-quality reconstruction, they can only divide the scene into static/dynamic parts and are unable to identify motion patterns.
A relevant recent work is Watch-It-Move~\cite{noguchi2021watch}, which achieves high-quality part-level reconstruction from images sequence.
However, it requires dense multi-view input and imposes ellipsoid-like priors to the part geometry, which may completely fail on challenging monocular data with complex scene geometry.
In contrast, our NeRF-based method does not require any shape priors of dynamic objects in complex scenes and can achieve dynamic reconstruction and part discovery from monocular input.

%% file: sections/prelim.tex
\section{Preliminaries: NeRF and D-NeRF}
By incorporating implicit function and volume rendering, Neural Radiance Field (NeRF)~\cite{mildenhall2020nerf} allows for scene reconstruction and novel view synthesis via optimizing scene representation directly.
In general, NeRF interprets static scenario as a continuous implicit function $\nerffunc$.
By querying spatial coordinates ($\coord$) and view direction ($\viewdir$), $\nerffunc$ outputs the corresponding density ($\density$) and observed color ($\pcolor$) as:
\begin{equation}
    (\pcolor,\density) = \nerffunc(\coord, \viewdir).
\end{equation}

Through classical volume rendering in graphics, the 3D scene representation $\nerffunc$ can be rendered into a 2D image.
Specifically, given a ray $\ray$ emitted from the optical center to a specific pixel in the image, the rendered color of that pixel is an integral of all the colors on the ray with near and far bounds $h_n$ and $h_f$:
\begin{equation}
    \label{equ:volumerender}
    \begin{split}
        \rcolor(\ray)=\int_{h_n}^{h_f} {T(h) \density\big(\ray(h)\big) \pcolor\big(\ray(h), \viewdir\big) dh},
        \quad \\
        \text{where} \quad
        T(h)=\exp\Big(-\int_{h_n}^{h}{\density(\ray(s)) ds}\Big). 
    \end{split}
\end{equation}
$T(h)$ can be interpreted as the transparency accumulated from $h_n$ to $h$.
Because of the inherent differentiability of Eq. \ref{equ:volumerender}, it only requires a set of images with camera poses to optimize $\nerffunc$ directly.

D-NeRF~\cite{pumarola2020d} extends NeRF to capture dynamic scenes, assuming that there is a static canonical space and includes all objects.
It divides the dynamic scene reconstruction in world space into two sub-problems: the NeRF representation learning of canonical space and the learning of the mapping from the world space to the canonical space (scene flow prediction) as:
\begin{equation}
    (\pcolor,\density) = \nerffunc(\Psi(\coord,t), \viewdir)
\end{equation}
where $\Psi(\coord,t)$ predict the position in the canonical space from $\coord$ at time $t$ into its canonical configuration.


%% file: sections/method.tex
\section{Our Method}
\begin{figure*}[t]
  \centering
  \includegraphics[width=\textwidth]{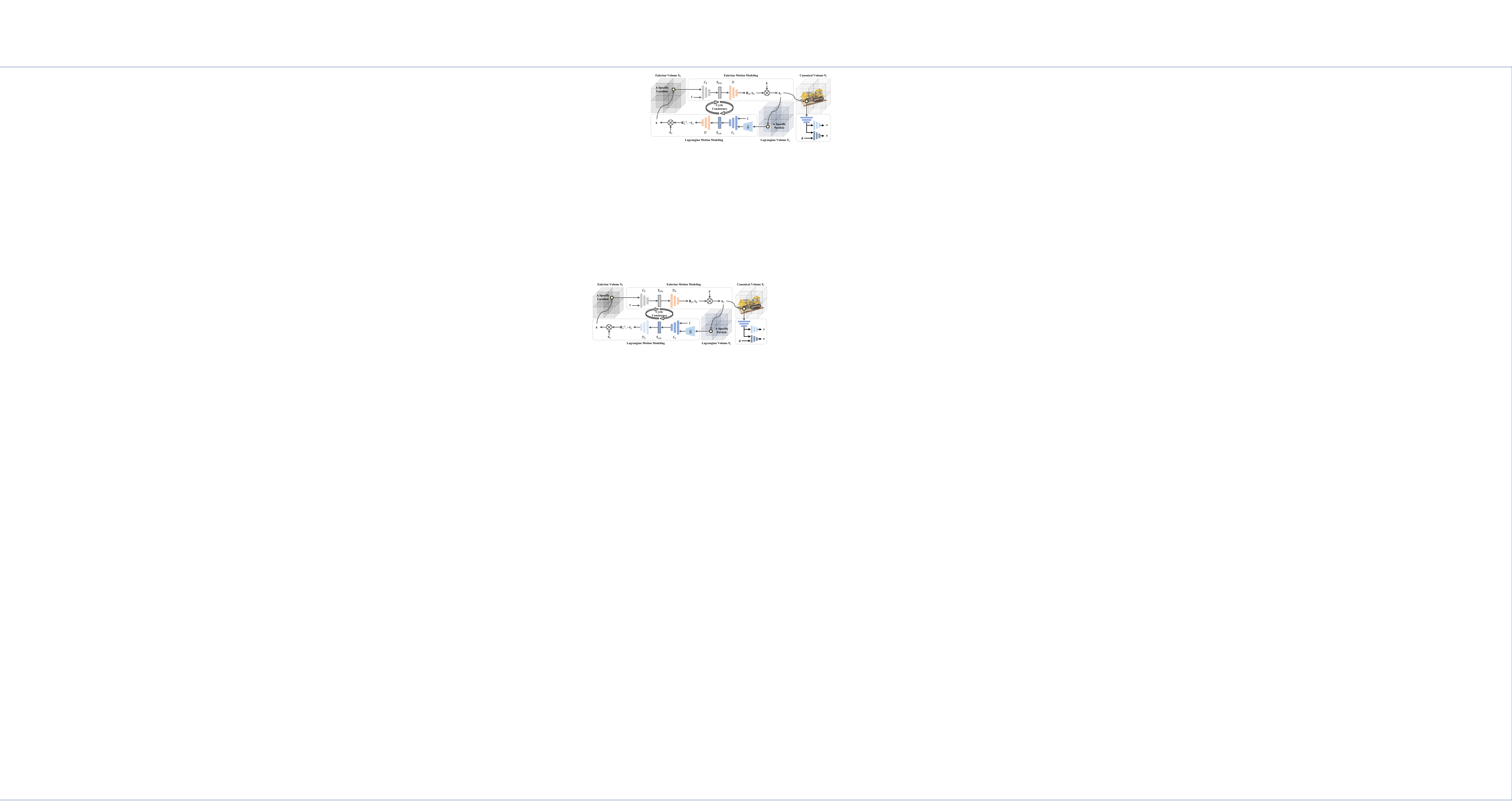}
  \caption{Overview of our method. Inspired by the Eulerian and Lagrangian viewpoints in fluid simulation, we design three modules to discover motion parts in a scene. The Eulerian module and the Lagrangian module observe the motion of specific spatial locations and specific particles, respectively. They both comprise a mutual mapping of a point between its position at an arbitrary time instance and its canonical configuration. The canonical module serves to reconstruct the geometry and appearance for volume rendering. Based on the particle trajectories recorded by the Lagrangian module, we can analyze the motion patterns and discover rigid parts.
   \label{fig:network}
  }
\end{figure*}

From the perspective of fluid simulation, the scene motion is composed of particle motions.
$\Psi(\coord,t)$ in D-NeRF can be interpreted as recording the motion of particles passing through a given coordinate $\coord$ at time $t$, corresponding to the Eulerian perspective.
Following D-NeRF, we also assume a canonical space that is static and includes all objects.
However, we describe the dynamic scene not only from the Eulerian perspective but also from the Lagrangian perspective.
Accordingly, we construct three modules, as Figure~\ref{fig:network} shows, that include an Eulerian module $\Psi_{E}(\coord,t)$ which maps a position $\coord$ at any time $t$ in the world space to the canonical space,
a Lagrangian module $\Psi_{L}(\coord_c,t)$ which tracks the trajectory of a particle corresponding to $\coord_c$ in the canonical space, and a canonical module which encodes the appearance and geometry in the canonical scene.
Under the assumption of finite rigid bodies, we exploit the learned motion by the Lagrangian module and design a motion grouping module to discover moving parts.
The particles in the same group share a common rigid transformation and should belong to the same part.
Next, we will describe these modules and loss functions in detail.

\subsection{Canonical Module}
\label{sec:canonicalmodule}
Same as NeRF, the canonical module is formulated as an implicit function $\nerffunc(\coord_c, \viewdir) \rightarrow (\pcolor,\density)$ which encodes the geometry and appearance in a canonical space.
To accelerate convergence, a hybrid representation of feature volume and neural network is adopted.
The queried canonical coordinate $\coord_c$ is first used to interpolate the corresponding features within a 3D feature volume $\mathbf{V}_c \in \mathbb{R}^{N_x \times N_y \times N_z \times C}$, where the $N_x \times N_y \times N_z$ denotes the spatial resolution and $C$ is the feature dimension.
To alleviate the local gradient artifact of grid representation, we adopt multi-distance interpolation and concatenate the features in different resolutions as~\cite{tineuvox}:
\begin{equation}
    \vf_{c} = \text{Tri-Interp}(\coord_c,\mathbf{V}) \oplus ... \oplus \text{Tri-Interp}(\coord_c,\mathbf{V}[::s_M]).
\end{equation}
After positional encoding, the queried feature with $\viewdir$ is fed into MLPs to predict $\density$ and $\pcolor$.

\subsection{Eulerian Module}
The Eulerian module $\Psi_{E}(\coord,t)$ records which particle $\coord_c$ in the canonical space goes through a specific location $\coord$ at the query time $t$.
Assuming that the scene is piece-wise rigid, we formulate this mapping as a rigid transformation in $\mathbb{SE}(3)$, ensuring that all the points on the same rigid body can be transformed using the same set of parameters.

Specifically, our Eulerian module contains three components.
First, the 3D feature volume $\mathbf{V}_{E}$ stores the information about the particles that pass through each position during the entire observation period.
Second, a motion extractor $\mathcal{E}_{E}$ decodes the motion feature from the interpolated feature in $\mathbf{V}_{E}$ at query time $t$.
Third, the rigid transformation decoder $\mathcal{D}_{E}$ maps the motion feature to rotation and translation parameters. 
The overall process can be formulated as:
\begin{equation}
\begin{aligned}
    \vf_{Em} = \mathcal{E}_{E}\left(\text{Tri-Interp}\left(\coord,\mathbf{V}_{E}\right), t \right)& \\
    (\mathbf{R}_{E}, \mathbf{t}_{E}) = \mathcal{D}_{E}(\vf_{Em})&
\end{aligned}
\end{equation}


Here we employ the continuous 6D intermediate representation~\cite{Zhou_2019_CVPR} for 3D rotation $\mathbf{R}_{E}$.
The Eulerian mapping from the world space at each temporal frame to the canonical space can be calculated by:
\begin{equation}
    \coord_c = \mathbf{R}_{E} (\coord-\mathbf{t}_{E})
\end{equation}

\begin{figure}[h]
  \centering
  \includegraphics[width=\linewidth]{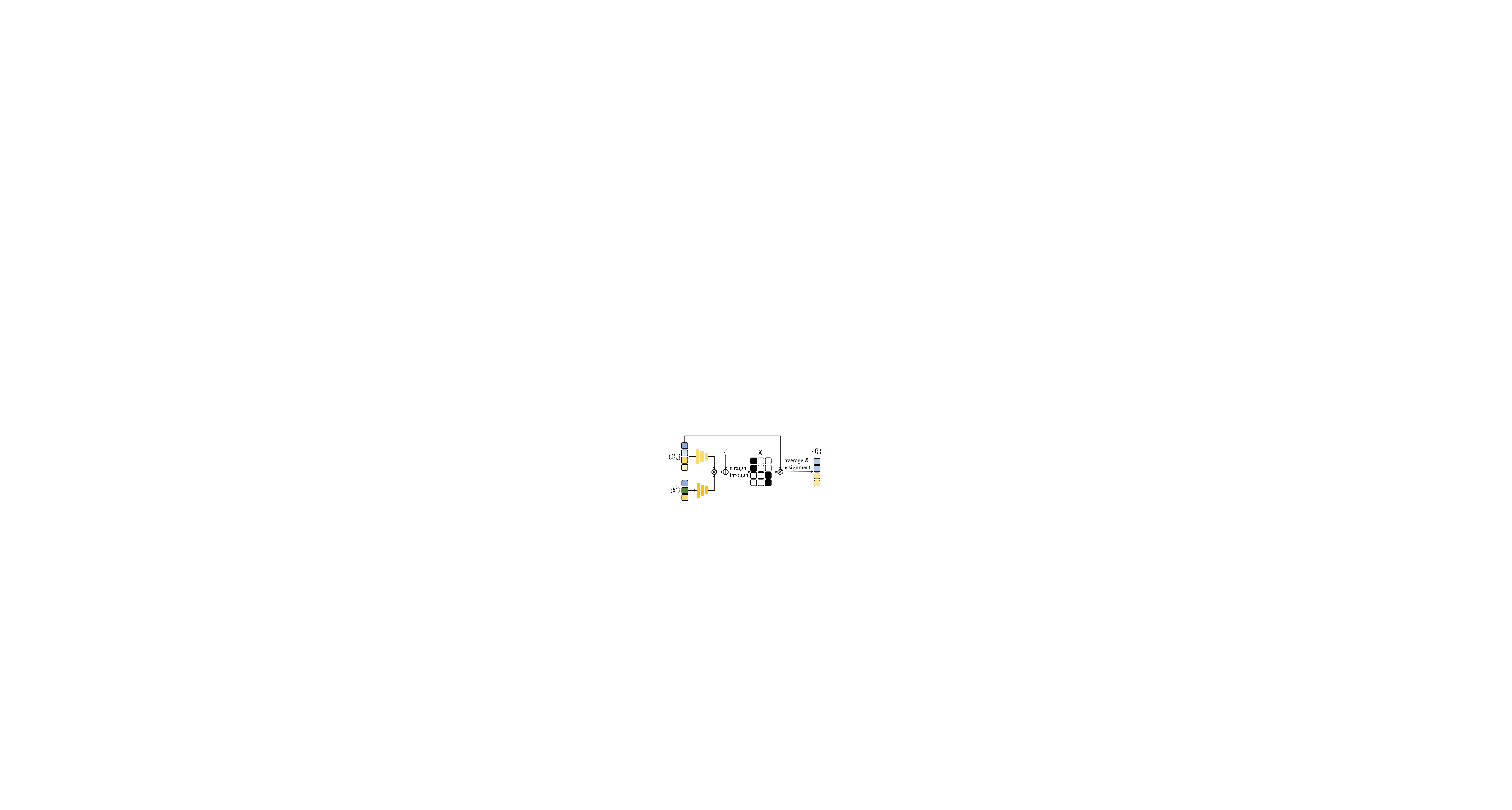}
  \caption{The architecture of our motion grouping network $\mathcal{G}$. First, we calculate the similarity between each fused feature $\vf_{L\coord}^i$ and each learnable slot $\mathbf{S}^l$, then the similarity-based hard grouping $\hat{\mathbf{A}}$ is implemented via the Gumbel-softmax and straight-through estimator trick. Finally, we assign $\Hat{\vf}_L^i$ to be the average of $\{\vf_L^i\}$ in its corresponding group by Equation~\ref{equ:average}.}
  \label{fig:grouping}
  \vspace{-3mm}
\end{figure}

\subsection{Lagrangian Module} 
As the inverse of the Eulerian module, the Lagrangian module $\Psi_{L}(\coord_c,t)$ tracks the trajectories of specific object particles over time.
We use the same manner to construct $\mathbf{V}_{L}$, $\mathcal{E}_{L}$ and $\mathcal{D}_{L}$.
Different from the Eulerian perspective, the trajectories of each particle in the Lagrangian perspective can be an important cue for rigid part discovery.
All particles belonging to the same rigid part share the same rigid body transformation at any time, which means that their motion can be represented by a single feature vector.
So we add an additional motion grouping network $\mathcal{G}$ after $\mathbf{V}_{L}$ to restrict that particle trajectories are only subject to a finite number of rigid motion patterns.

Similar to~\cite{xu2022groupvit}, we use the attention module with the straight-through estimator trick to achieve the hard grouping of Lagrangian view features.
To encourage the spatial coherence of points in the same group, the coordinate of each point $\coord_c^i$ is concatenated to the corresponding Lagrangian feature $\vf_{L}^i$.
Specifically, we first compute the similarity map $\mathbf{A}$ between the feature $\{\vf_{L\coord}^i = \vf_{L}^i \oplus \coord_c^i \}$ and learnable slots $\{\mathbf{S}^l\}$ by Gumbel-softmax:
\begin{equation}
    \mathbf{A}_{il} = \frac{\text{exp} (\vb{W}_q \vf_{L\coord}^i \cdot \vb{W}_k \mathbf{S}^l + \gamma_i)}
    {\sum_{1}^{L} \text{exp} (\vb{W}_q \vf_{L\coord}^i \cdot \vb{W}_k \mathbf{S}^l + \gamma_i)},
\end{equation}
where $\vb{W}_q$ and $\vb{W}_k$ are linear mappings and $\gamma$ is a sample drawn from $Gumbel(0,1)$. 
Then the straight-through estimator trick is used to convert the soft similarity map to one-hot formulation:
\begin{equation}
\label{equ:straight-through}
    \hat{\mathbf{A}} = \text{one-hot}(\mathbf{A}_{\text{argmax}}) + \mathbf{A} - \text{detach}(\mathbf{A})
\end{equation}
where the detach operation cuts off the corresponding gradient.
Despite the hard conversion, Eq.~\eqref{equ:straight-through} can still keep the gradient the same as $\mathbf{A}$.
The hard similarity matrix $\hat{\mathbf{A}}$ distributes all the Lagrangian features into several groups, where each group represents the particles with the same motion pattern.
Instead of directly assigning the learned slot as the updated Lagrangian feature, we also calculate the average of all the Lagrangian features in the same group, and use them to update the original Lagrangian features. In this way, each updated Lagrangian feature will be directly related to the Lagrangian grid $\mathbf{V}_{L}$, allowing for more efficient optimization. This procedure can be formulated as:
\begin{equation}
\label{equ:average}
    \Hat{\vf}_L^i = 
    \frac{
        \sum_{1}^{I} \hat{\mathbf{A}}_{il} \cdot \vf_L^i
        }
    {\sum_{1}^{I} \hat{\mathbf{A}}_{il}}
\end{equation}
Then the updated Lagrangian features $\Hat{\vf}_L^i$ with query time $t$ is fed into $\mathcal{E}_{L}$ and $\mathcal{D}_{L}$ sequentially to decode the motion feature $\vf_{Lm}$ and the rigid transformation $\vb{R}_L$, $\vb{t}_L$.
As mentioned in Section~\ref{sec:loss}, to efficiently implement the cycle consistency between the Eulerian and Lagrangian modules, we expect $\vb{R}_L = \vb{R}_E$ and $\vb{t}_L = \vb{t}_E$.
So the Lagrangian mapping from the canonical space to the world space at each temporal frame is calculated by:
\begin{equation}
    \coord = \vb{R}_{L}^{-1} (\coord_c + \vb{t}_{L})
\end{equation}

\subsection{Loss Functions}
\label{sec:loss}
As our main optimization goal, we adopt the Mean Squared Error (MSE) between the rendered pixel color and the ground truth pixel color as our reconstruction loss: 
\begin{equation}
    \mathcal{L}_{\text{photo}} = \frac{1}{|\mathcal{R}|}
    \sum_{\ray \in \mathcal{R}} \|\hat{\rcolor}(\ray)-\rcolor(\ray)\|^2_2.
\end{equation}
Also, following~\cite{Sun_2022_CVPR}, the per-point color loss $\mathcal{L}_{\text{per\_pt}}$ and background entropy loss $\mathcal{L}_{\text{entropy}}$ are used to directly supervise the sampled point color and encourage the background probability to concentrate around 0 or 1.

In addition, a cyclic consistency loss is designed to encourage the reciprocity of the Lagrangian module and the Eulerian module.
Instead of measuring the difference between rigid transformation $(\vR_{L}, \vt_{L})$ and $(\vR_{E}, \vt_{E})$, we found that accounting for the difference between low-level motion features $\vf_{Lm}$ and $\vf_{Em}$ leads to more robust optimization. The cycle loss is defined as
\begin{equation}
    \mathcal{L}_{\text{cycle}} = \frac{1}{|\mathcal{P}_{obj}|}
    \sum_{\coord \in \mathcal{P}_{obj}} \|\vf_{Lm}^\coord - \vf_{Em}^\coord\|^2_2.
\end{equation}
Since the deformation of free space does not satisfy the assumption of finite rigid motions, $\mathcal{L}_{\text{cycle}}$ is only calculated at sampled points on objects $\{ \coord \in \mathcal{P}_{obj} | \density_\coord > \epsilon \}$. In our experiments, $\epsilon = 10^{-4}$. 
The overall loss function is:
\begin{equation}
\begin{split}
\mathcal{L} = \mathcal{L}_{\text{photo}} + 
              w_{\text{cycle}}\mathcal{L}_{\text{cycle}} +
              w_{\text{per\_pt}} \mathcal{L}_{\text{per\_pt}} + \\
              w_{\text{entropy}} \mathcal{L}_{\text{entropy}} + 
              w_{\text{tv}}  \mathcal{L}_{\text{tv}}.
\end{split}
\end{equation}
\begin{figure}[h]
  \centering
  \includegraphics[width=\linewidth]{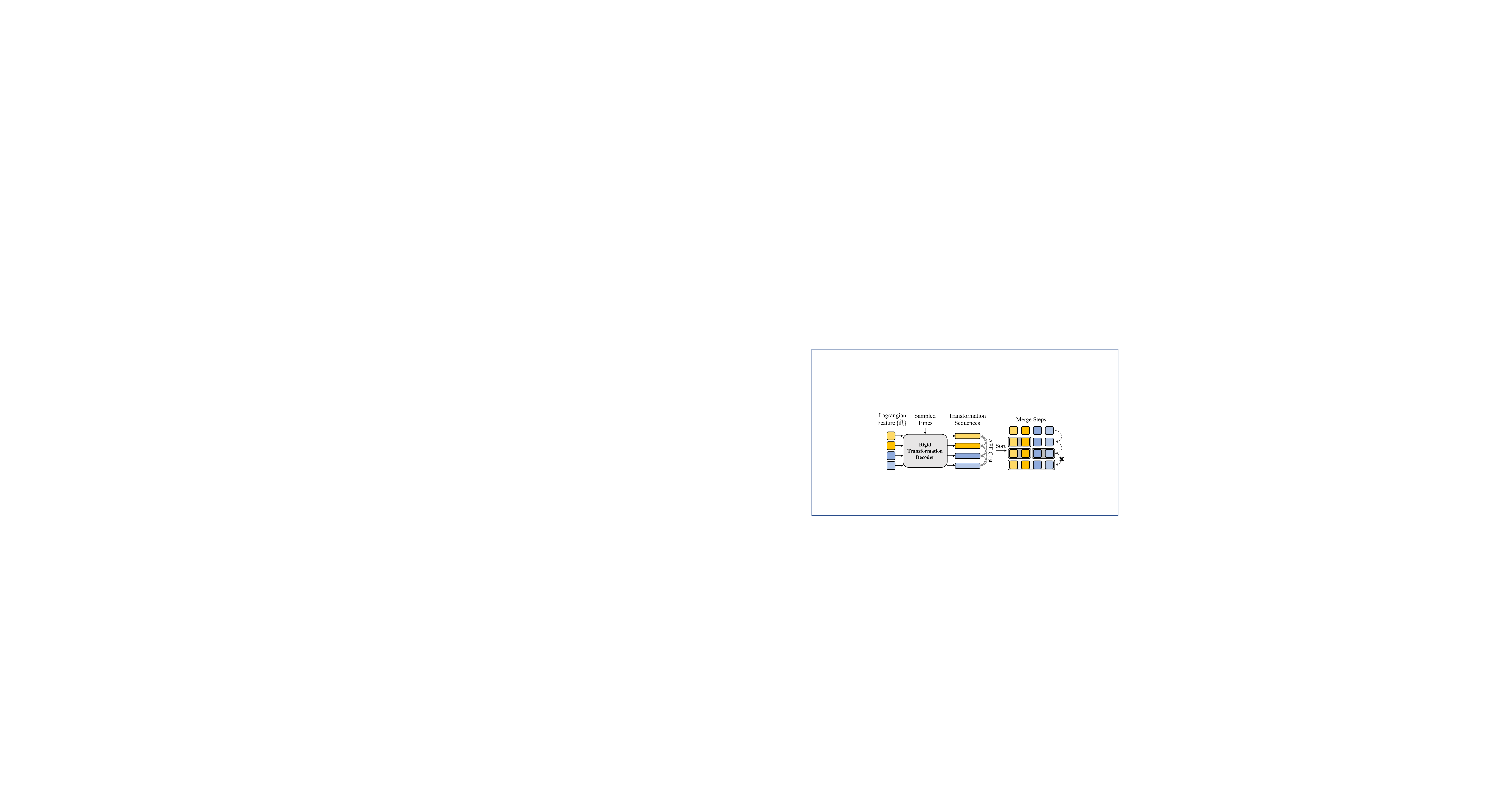}
  \caption{The group merging procedure. We decode the average Lagrangian features of each group into rigid transformation sequences and determine the merge order as well as the stop step by evaluating the APE cost between the sequences.}
  \label{fig:merge}
  \vspace{-2mm}
\end{figure}

\begin{figure*}[t]
  \centering
  \includegraphics[width=\textwidth]{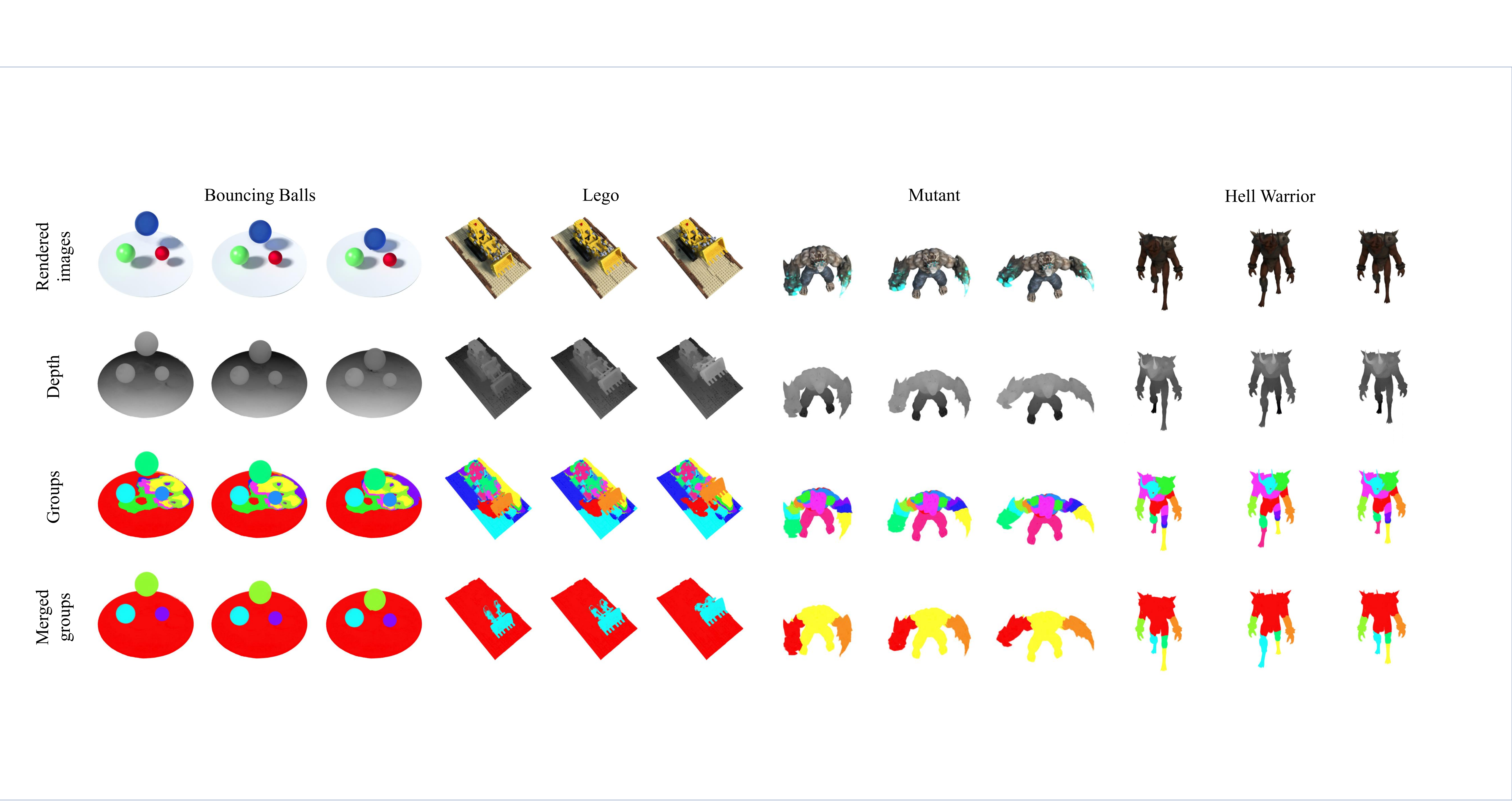}
  \caption{Visualization of our results on the D-NeRF synthetic dataset. The first two rows show the image and depth rendering results at different timesteps. The third row shows our initial grouping results. We assign a specific color to each group in this visualization. The last row shows the grouping results after merging, and it shows that our merging module could adaptively aggregate groups with similar motions and achieve clean and accurate segmentation results. 
  }
  \label{fig:dnerf}
  \vspace{-3mm}
\end{figure*}

\subsection{Group Merging Module} 
\label{sec:merge}
It is not reasonable to use the same number of groups for all scenarios.
We generally set a large number of groups as an upper bound on the number of rigid bodies in the scene.
We observe that points with the same motion pattern may distribute in different groups, causing over-segmented results~(Figure \ref{fig:dnerf}). 
This is because we provide an excessive number of groups and the same rigid transformations could be easily represented by very different high-level motion features.
To this end, we design an efficient heuristic algorithm for group merging based on motion differences.
1) We sample points uniformly in canonical space and filter the free space points by removing points with density lower than the threshold $\epsilon$.
2) These remained points are fed into the Lagrangian module to get the updated feature $\Hat{\vf}_L^i$, which is the high-level representation of each motion group.
3) We evaluate the rigid transformation similarity between each pair of groups:
The rigid transformation sequences are generated by decoding the updated slots into rotation and translation with uniformly sampled times between 0 and 1.
4) We use the Absolute Pose Error (APE) to measure the difference between each sequence pair:
\begin{equation}
    APE_{i,j} = \sum_{t} \|(\mathbf{P}_{i}^{t})^{-1} \mathbf{P}_{j}^{t} - \mathbf{I}_{4\times4}\|,
\end{equation}
where $\mathbf{P}_{i}^{t}$ is the transformation matrix of group $i$ at time $t$.
5) We recursively find the two groups with the smallest APE at the current step and record their merge APE cost until all the groups are merged into a single one.
In the early stages, the groups with similar motion patterns are merged, which keeps the merging cost growth slow.
Once groups representing different motions are merged, the cost will jump, indicating that the merging process should terminate.
In practice, we simply find the step with the largest cost increase compared to the previous step and choose its previous step as the final result.


%% file: sections/experiments.tex
\section{Experiments and Results}
Our method not only enables high-quality dynamic scene reconstruction but also allows for the discovery of reasonable rigid parts.
In this section, we first evaluate the reconstruction and part discovery performance of our method on the D-NeRF $360^{\circ}$ synthetic dataset.
Then, we construct a synthetic dataset with ground-truth motion masks to quantitatively evaluate our motion grouping results.
Finally, we provide direct applications of our method for scene editing.

\subsection{Implementation}
We use $50 \times 50 \times 50$ voxels to construct the Eulerian volume $\mathbf{V}_E$ and the Lagrangian volume $\mathbf{V}_L$ with a feature dimension of $20$.
The canonical volume $\mathbf{V}_c$ is constructed with a $160 \times 160 \times 160$ voxel.
Following~\cite{tineuvox}, the feature dimension of $\mathbf{V}_c$ is set as 6.
To alleviate the optimization difficulty and speed up training, we set the initial resolution of the canonical volume to $40^3$ and upsample it at the $4k$, $6k$, and $8k$ iterations.
For the MLPs in our framework, we set the channel number of all hidden layers to 128.
We use two-layer MLPs for the motion extractors $\mathcal{E}_{E}$ and $\mathcal{E}_{L}$.
Two separated linear layers are used to predict the 6D rotation and 3D translation with biases as $(1, 0, 0, 0, 1, 0)$ and $(0,0,0)$, respectively, so that the initial deformation is an identity.
For motion grouping, we set the slot number to 12 and initialize the slots from a standard normal distribution.
We use the Adam optimizer for a total of $20k$ iterations, 
by sampling 4096 rays from a randomly sampled image in each iteration.
To reduce the learning difficulty, we add images into the training set progressively at the early training stage.
We set the learning rate as 0.08 for the Eulerian and Lagrangian volumes, 0.01 for the canonical volume, $6 \times 10^{-4}$ for $\mathcal{E}$ and $\mathcal{D}$, $8 \times 10^{-4}$ for other networks.
All the experiments were conducted on a single NVIDIA RTX3090 GPU.

\begin{table*}[t]
	\caption{Comparison with NeRF-based methods. We indicate the best and second best with bold and underlined respectively. We not only achieve high-quality realistic rendering results on par with state-of-the-art method (the concurrent TiNeuVox) but also discovers reasonable rigid parts. Our method is highly efficient and can achieve both tasks in about 26 minutes. }
	\label{tb:dnerf}
        \vspace{-2mm}
	\begin{center}
		\begin{tabular}{l|cc|ccccccccc}
                \specialrule{0.07em}{0pt}{0pt}
    		\multirow{2}*{Method} &
    		\multirow{2}*{Part} &
    		\multirow{2}*{$T_{train}$} &
    	    \multicolumn{2}{c}{Hell Warrior} &
    		\multicolumn{2}{c}{Mutant} &
    		\multicolumn{2}{c}{Hook} &
    		\multicolumn{2}{c}{Bouncing Balls} \\
    		
                &  &  &
    		PSNR  &
    		SSIM  &
    		PSNR   &
    		SSIM   &
    		PSNR   &
    		SSIM   &
    		PSNR   &
    		SSIM  &\\
      
            \hline
            
			T-NeRF & $\usym{2717}$ & $\sim$ hours &
			23.19 & 0.93 &
			30.56 & 0.96 &
			27.21 & 0.94 &
			37.81 & \underline{0.98} \\
			
			D-NeRF & $\usym{2717}$ & 20 hours &
			25.02 & \underline{0.95} & 
			31.29 & 0.97 & 
			29.25 & \underline{0.96} & 
			38.93 & \underline{0.98} \\
			
			TiNeuVox & $\usym{2717}$ & 28 mins &
			\underline{28.17} & \textbf{0.97} & 
			33.61 & \underline{0.98} & 
			\textbf{34.45} & \textbf{0.97} & 
			\textbf{40.73} & \textbf{0.99} \\
			
			NDVG & $\usym{2717}$ & \textbf{23 mins} & 
			25.53 & \underline{0.95} & 
			$\textbf{35.53}$ & $\textbf{0.99}$ & 
			29.80 & \underline{0.96} & 
			34.58 & 0.97 \\
			
			WIM & $\usym{1F5F8}$ & $\sim$ hours &
			12.35 & 0.81 &
			16.20 & 0.85 & 
			14.16 & 0.82 & 
			15.82 & 0.84 \\
			
			Ours & $\usym{1F5F8}$ & \underline{26 mins} &
			$\textbf{28.66}$ & $\textbf{0.97}$ & 
			\underline{34.42} & \underline{0.98} & 
			\underline{31.39} & $\textbf{0.97}$ & 
			\underline{38.99} & $\textbf{0.99}$ \\
			\specialrule{0.0em}{0.5pt}{0pt}
   
			\specialrule{0.07em}{0pt}{0pt}
			
    		\multirow{2}*{Method} &
    
    		\multicolumn{2}{c|}{Average} &
    		\multicolumn{2}{c}{Lego} &
    		\multicolumn{2}{c}{T-Rex} &
    		\multicolumn{2}{c}{Stand Up} &
    		\multicolumn{2}{c}{Jumping Jacks} \\
    		
    		~ & PSNR & SSIM &
    		PSNR  &
    		SSIM   &
    		PSNR   &
    		SSIM  &
    		PSNR  &
    		SSIM  &
    		PSNR   &
    		SSIM  &\\

		    \hline
                
			T-NeRF & 29.50 & 0.95 &
			23.82 & 0.90 &
			30.19 & 0.96 &
			31.24 & 0.97 &
			32.01 & \underline{0.97} \\
			
			D-NeRF & 30.43 & 0.95 &
			21.64 & 0.83 & 
			31.75 & \underline{0.97} & 
			32.79 & \underline{0.98} & 
			\underline{32.80} &\textbf{0.98} \\
			
			TiNeuVox & \textbf{32.67} & \textbf{0.97} & 
			25.02 & \underline{0.92} & 
			$\textbf{32.70}$ & $\textbf{0.98}$ & 
			$\textbf{35.43}$ & $\textbf{0.99}$ & 
			$\textbf{34.23}$ & $\textbf{0.98}$ \\
			
			NDVG & 30.54 & \underline{0.96} & 
			$\textbf{25.23}$ & $\textbf{0.93}$ & 
			30.15 & \underline{0.97} & 
			34.05 & \underline{0.98} & 
			29.45 & 0.96 \\
			
			WIM & 15.72 & 0.83  &
			13.95 & 0.72 & 
			19.05 & 0.87 & 
			16.26 & 0.89 & 
			17.95 & 0.87 \\
			
			Ours & \underline{32.24} & \textbf{0.97} &
			\underline{25.08} & \underline{0.92} & 
			\underline{32.24} & \textbf{0.98} & 
			\underline{34.46} & \underline{0.98} & 
			32.22 & \textbf{0.98} \\

                \specialrule{0.0em}{0.5pt}{0pt}
			\bottomrule
		
		\end{tabular}
	\end{center}

\vspace{-6mm}
\end{table*}

\subsection{Evaluation on D-NeRF dataset}
We adopt the 360$^\circ$ Synthetic dataset provided by D-NeRF~\cite{pumarola2020d} to evaluate our method quantitatively and qualitatively. 
The dataset contains eight synthetic dynamic scenes with different motion patterns, and only one view is captured at each time step.
We compare our method with the state-of-the-art Eulerian-view methods: D-NeRF~\cite{pumarola2020d}, TiNeuVox~\cite{tineuvox}, NDVG~\cite{Guo_2022_NDVG_ACCV}, and a Lagrangian-view method WIM~\cite{noguchi2021watch}.
For TiNeuVox, we use their base version with a canonical grid in $160^3$ resolution and hidden layers of 256 channels.
Following these previous works, we train each scene with images at $400 \times 400$ resolution and use three metrics for evaluation: Training time ($T_{train}$), Peak Signal-to-Noise Ratio (PSNR), and Structural Similarity (SSIM).

As shown in Table~\ref{tb:dnerf}, while keeping the training time within 30 minutes on one GPU, our method not only achieves high rendering quality but also supports part discovery.
Compared to the previous methods, we achieved the best and second best in SSIM and PSNR, respectively.
Compare to TiNeuVox, our method has a slight PSNR drop.
The main reason is that TiNeuVox employs a temporal enhancement module in the canonical space to improve quality, which also leads to a time-varying canonical space.
After removing this enhancement module in TiNeuVox, its average PSNR drops to 31.47.
In our paper, to achieve better disentanglement of geometry and motion, we expect the geometric evolution only comes from the scene motion. 
Therefore we did not adopt a similar enhancement strategy to form a time-invariant canonical space.
For WIM, due to the nonexistence of canonical space, the significant motion ambiguity under the single view setting causes the failure.

We show our visualization results in Figure~\ref{fig:dnerf}.
It can be seen that our method enables high-quality appearance and geometry reconstruction.
We also assign each query point the corresponding group color and render it to 2D images.
As discussed in Section~\ref{sec:merge}, over-segmentation occurs because similar motion could be represented by different high-level features (see the third row in Figure~\ref{fig:dnerf}).
Through our group merging algorithm, we only retain the highly distinguishable motion modes and obtain concise part segmentation.
Thanks to the motion-based grouping mechanism, our method is capable of overlooking motion-irrelevant characteristics in geometry and appearance and producing clean part discovery results on these realistic complex scenes.



\subsection{Motion Grouping Evaluation}
\begin{figure}[t]
  \centering
  \includegraphics[width=\linewidth]{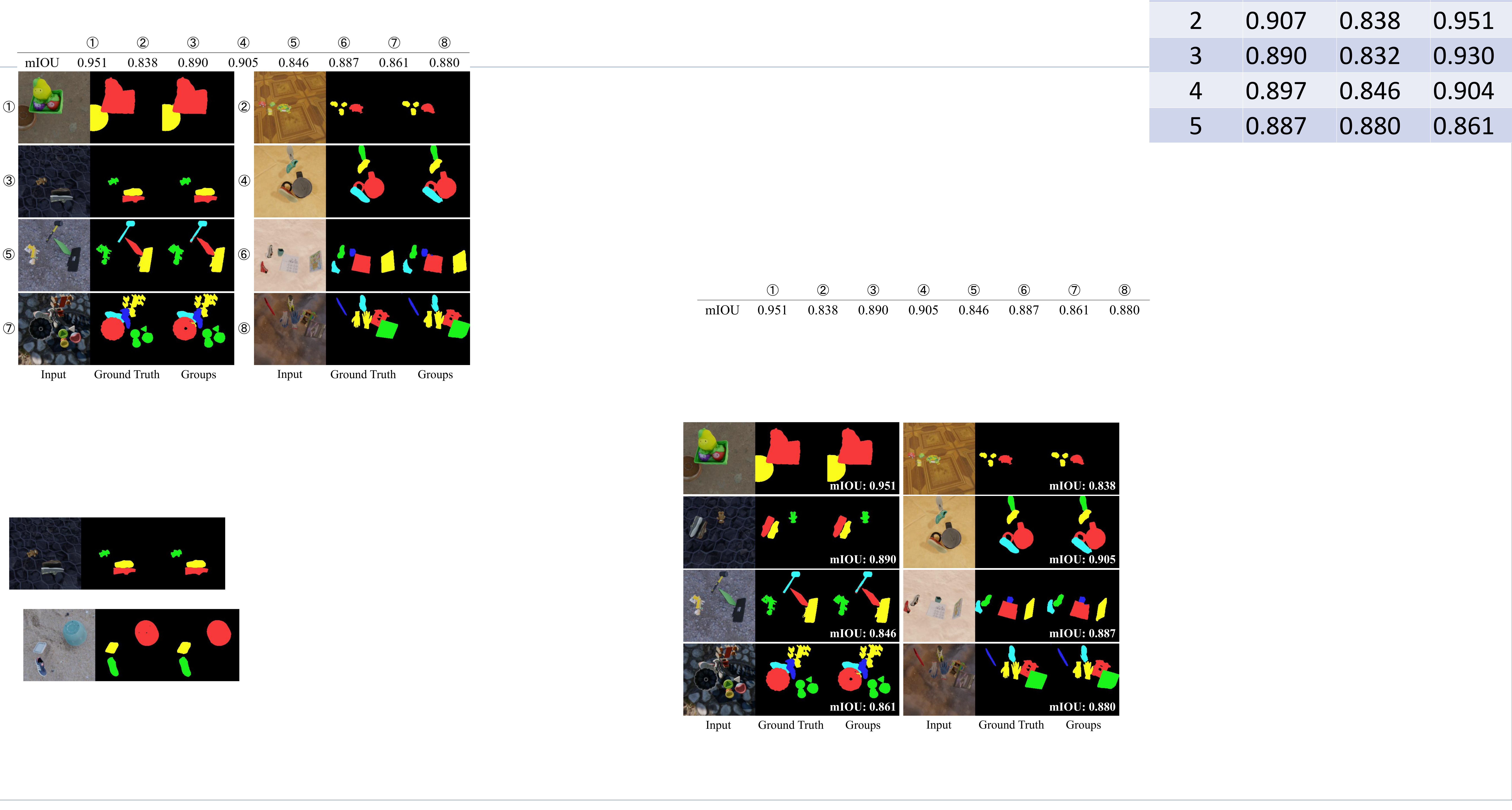}
  \caption{Motion grouping evaluation on our generated datasets. Our method is able to achieve high mIOU scores across various scene configurations and has demonstrated the ability to handle complex geometry and topology of different objects.
  }
  \vspace{-2mm}
  \label{fig:kubric}
\end{figure}

In this section, we provide a quantitative evaluation of our motion grouping results. 
We created a synthetic dataset with ground truth image-segmentation pairs using Kubric~\cite{greff2021kubric} toolkit.
Each scene contains 1 to 5 realistic real-world objects from the GSO dataset~\cite{ICRA46639} with different initial velocities.
We followed the same sampling and rendering process as D-NeRF~\cite{wu2022dnerf} to generate a 120-frame monocular image sequence with a resolution of 256$\times$256 for each scene.
To begin our evaluation, we first establish the correlation pairs between the ground truth label and our predicted groups. 
For each group, we assign the ground truth label with the highest number of pixels corresponding to it in the first 10 frames. More details are included in the supplementary material. 

We calculate the mean Intersection over Union (IOU) for the assigned label mask with its corresponding ground truth mask over the entire image sequence. 
It is noted that achieving a high mIOU score over the entire sequence requires more than just the ability to accurately distinguish each individual part. It also necessitates the capacity to consistently track each part throughout the sequence.

We present 8 examples in Figure~\ref{fig:kubric}, showcasing both quantitative mIOU and qualitative comparisons with the ground truth masks. 
Additional results can be found in the supplementary material. 
Despite the variation in the scene configurations, our method achieves an mIOU score of over $85\%$ on most scenes, clearly demonstrating its robustness over the dataset.
Moreover, the high mIOU score indicates that our method can generate accurate part segmentation results and continuously track specific parts throughout the sequences, ensuring both temporal and multi-view consistency of the discovered parts.
Furthermore, our method is capable of dealing with complex geometry and topology. Holes and geometry details are nicely revealed by our method-- see the cable (row 2, col 2) and bookshelf (row 4, col 2) examples in Figure~\ref{fig:kubric}. By utilizing our motion-based grouping approach, the method can accurately segment objects even if they are spatially separated-- see the 3-car (row 1 col 2) and gloves (row 4 col 2) examples in Figure~\ref{fig:kubric}.

\begin{figure}[t]
  \centering
  \includegraphics[width=\linewidth]{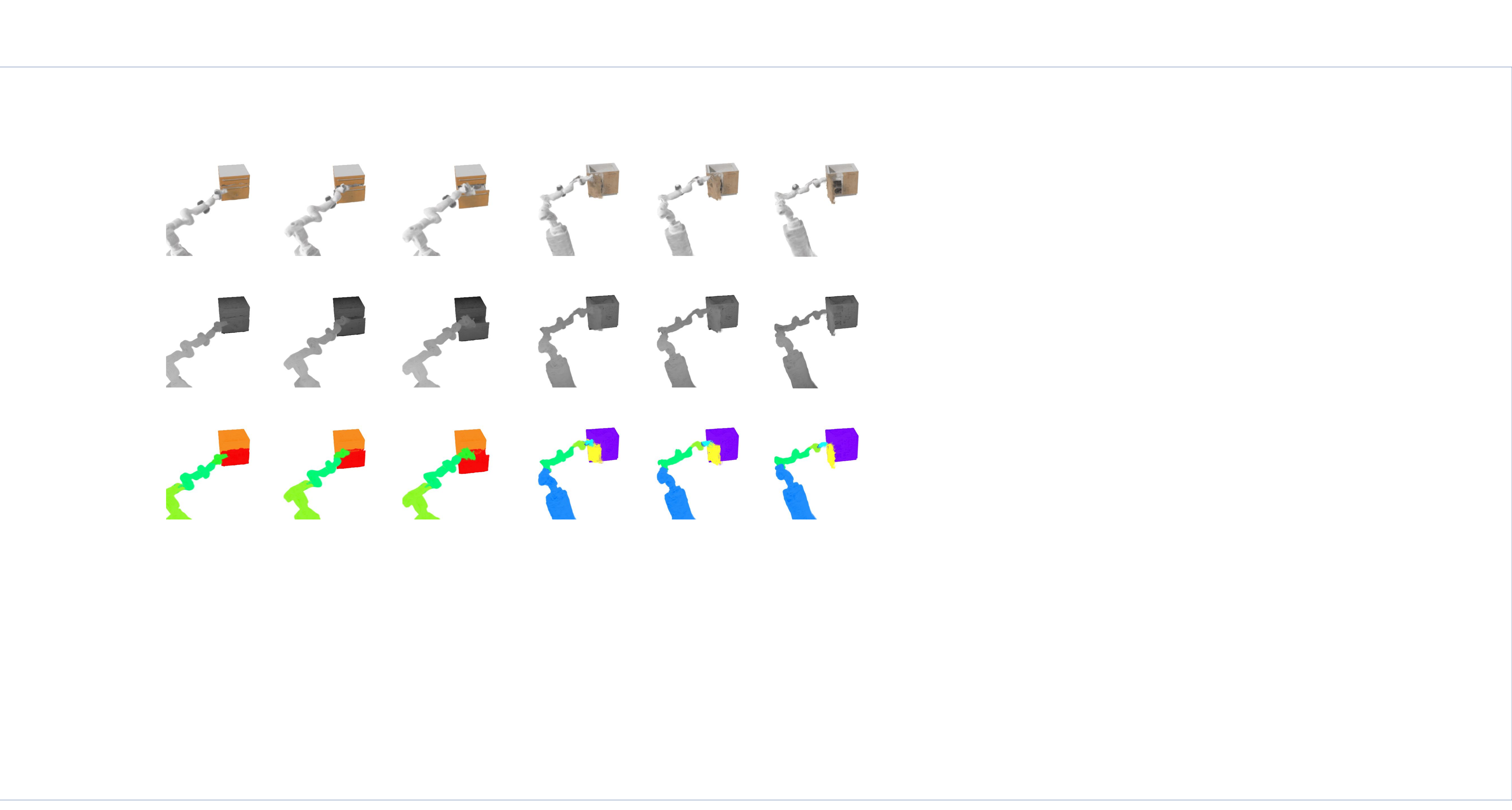}
  \caption{Visualization of our results on two robotic manipulation sequences. By observing the robot interacting with the object, we can discover dynamic parts with their trajectories, which is essential for functionality reasoning.
  }
  \label{fig:maniskill}
  \vspace{-2mm}
\end{figure}

\subsection{Application: Structured Scene Modeling by Robotic Manipulation}
\begin{figure}[t]
  \centering
  \includegraphics[width=\linewidth]{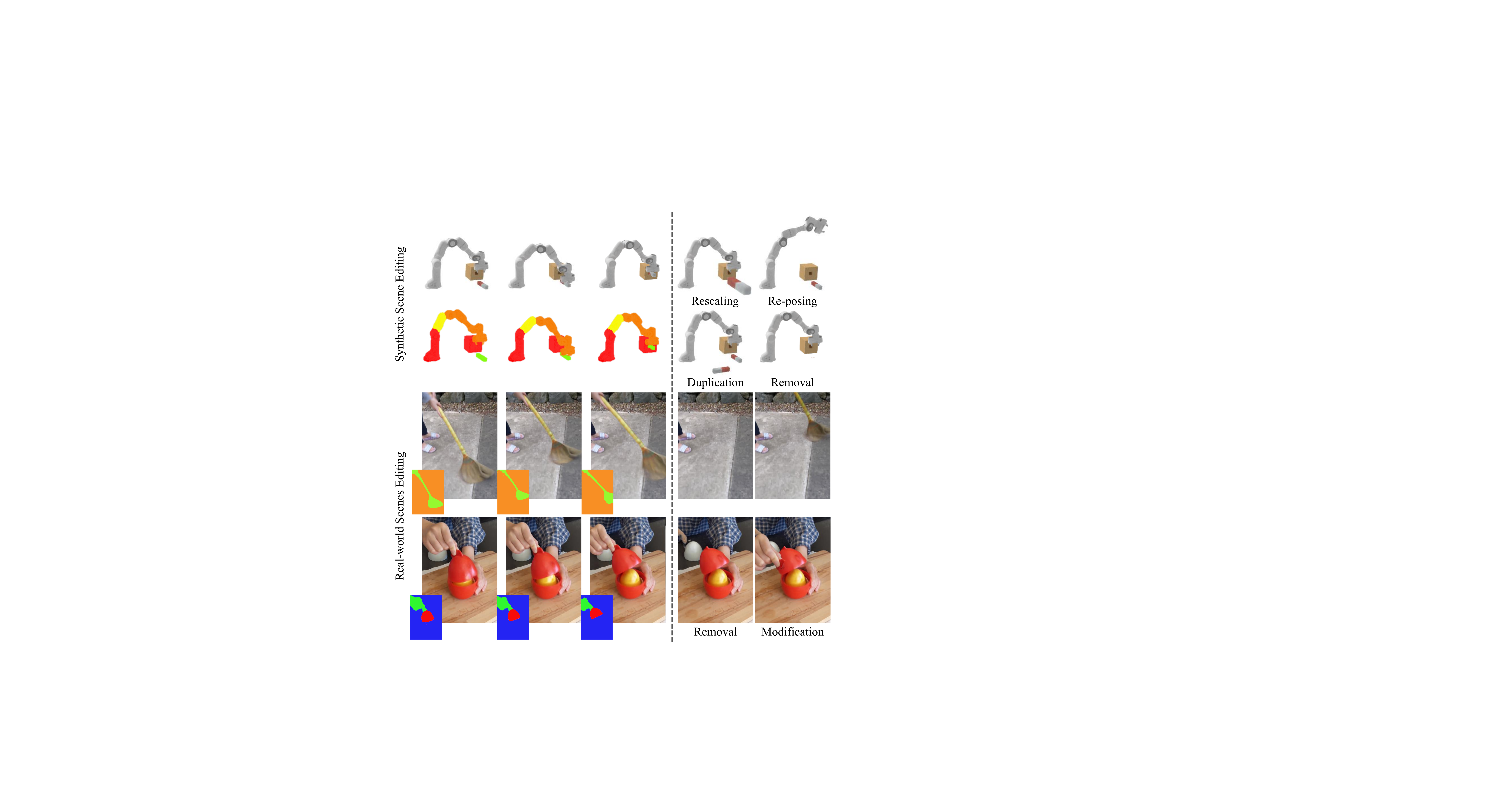}
  \caption{Scene editing on the synthetic data and real data. With the discovered parts, we can directly rescale, repose, duplicate, or remove specific parts, enabling us to easily create novel scenes.}
 \label{fig:editing}
 \vspace{-2mm}
\end{figure}

Observation and interaction are crucial for human beings to learn from the real world. 
In this section, we show that our method can identify objects and understand the functionality of their parts by observing physical interaction procedures. 
To demonstrate this, we utilized a subset of the robotic manipulation environments in ManiSkill~\cite{mu2021maniskill} to simulate robot-object interactions and use a similar monocular camera setting as~\cite{pumarola2020d}.
We conducted experiments using two different setups: rigid-body manipulation and articulated object manipulation. 
In the first setup, the robot is tasked with grasping a specific object and moving it to a target position, as illustrated in Figure~\ref{fig:editing}.
In the second setup, the robot performs operations on a particular movable part of an articulated object, as shown in Figure~\ref{fig:maniskill}.
In both setups, through observation, our method can accurately identify the manipulated object or object part, such as the drawer (orange) and the cabinet door (yellow) in Figure~\ref{fig:maniskill}, as well as the links and joints of the robot.
The discovered 3D parts with their Lagrangian motion could provide a strong prior for downstream functionality reasoning and robotic reinforcement learning tasks.


\subsection{Application: Scene Editing}
With the help of the Lagrangian module, we can obtain the structural representation of dynamic scenes, which can be applied to downstream tasks directly.
Figure~\ref{fig:editing} presents a few scene-editing applications supported by our approach in ManiSkill synthetic sequence~\cite{mu2021maniskill} and HyperNeRF real-world sequence~\cite{park2021hypernerf}.
Since our method conducts grouping in the 3D canonical space, the consistency can be maintained not only across multiple views but also across time steps.
In addition to understanding scenes, our method can also edit scenes and generate new renderings from the scene.
In the synthetic data, we show the duplication, removal, scaling of specific parts in a scene, and the re-posing of an articulated robot arm.
Furthermore, we demonstrate the scalability of our method to real-world scenarios by showcasing the removal or modification of the position of the broom and hand respectively in these two real scenes.

%% file: sections/discussion.tex
\section{Conclusion}
In this paper, we present MovingParts, a novel method for 3D dynamic scene reconstruction and part discovery.
Inspired by fluid simulation, we observe the motion in the scene from both the Eulerian view and the Lagrangian view.
In the particle-based Lagrangian view, we constrain the motion pattern of the particles to be a few rigid transformations, so that we successfully perform part discovery.
To ensure fast convergence during training, we utilize a hybrid feature volume and neural network representation, for both views which are efficiently supervised by a cycle-consistency loss.
What is more, the learned part representation could directly be applied to downstream tasks, e.g., object tracking, structured scene modeling, editing, etc.

\vspace{1em}
\noindent\textbf{Limitations.} 
Motion modeling at a specific location can be considered as a sequence decoding task. In this paper, we explicitly store the motion features in low-dimensional vectors, which makes it challenging to model motion on very long sequences. Although we can circumvent the issue by manually splitting long sequences into shorter ones, a unified long sequence encoding-decoding scheme will still be a more elegant and efficient solution. We defer the exploration of this challenging setting to future work.